\documentclass[10pt,twocolumn,letterpaper]{article}

\usepackage{cvpr}
\usepackage{times}
\usepackage{epsfig}
\usepackage{graphicx}
\usepackage{amsmath}
\usepackage{amssymb}

\usepackage{times}
\usepackage{epsfig}
\usepackage{graphicx}
\usepackage{amsmath}
\usepackage{amssymb}
\usepackage{multirow}
\usepackage{bbm}
\usepackage{times}
\usepackage{epsfig}
\usepackage{graphicx}
\usepackage{amsmath}
\usepackage{amssymb}
\usepackage[table]{xcolor}
\usepackage{multirow}
\usepackage[ruled,linesnumbered]{algorithm2e}
\usepackage{amsmath}  
\usepackage{bm}       

\usepackage{booktabs}
\def\bx{\mathbf{x}}
\def\bw{\mathbf{w}}
\def\bW{\mathbf{W}}
\def\bh{\mathbf{h}}

\def\bz{\mathbf{z}}

\def\bmu{\bm{\mu}}
\def\bphi{\bm{\phi}}
\def\btheta{\bm{\theta}}
\def\bsigma{\bm{\sigma}}
\def\bdelta{\bm{\delta}}

\DeclareMathOperator*{\sgn}{sgn}

\begin{document}

\title{Boosting Unconstrained Face Recognition with Targeted Style Adversary}

\author{Mohammad Saeed Ebrahimi Saadabadi, Sahar Rahimi Malakshan,\\
Seyed Rasoul Hosseini, and Nasser M. Nasrabadi\\
{\tt\small{me00018, sr00033, sh00111}@mix.wvu.edu, nasser.nasrabadi@mail.wvu.edu}
}

\maketitle
\thispagestyle{empty}

\begin{abstract}
While deep face recognition models have demonstrated remarkable performance, they often struggle on the inputs from domains beyond their training data.
Recent attempts aim to expand the training set by relying on computationally expensive and inherently challenging image-space augmentation of image generation modules.
In an orthogonal direction, we present a simple yet effective method to expand the training data by interpolating between instance-level feature statistics across labeled and unlabeled sets.
Our method, dubbed Targeted Style Adversary (TSA), is motivated by two observations: (i) the input domain is reflected in feature statistics, and (ii) face recognition model performance is influenced by style information. Shifting towards an unlabeled style implicitly synthesizes challenging training instances. We devise a recognizability metric to constraint our framework to preserve the inherent identity-related information of labeled instances.
The efficacy of our method is demonstrated through evaluations on unconstrained benchmarks, outperforming or being on par with its competitors while offering nearly a 70\% improvement in training speed and 40\% less memory consumption.
\end{abstract}

\section{Introduction}
\label{sec:intro}
Using large-scale training benchmarks, well-constructed Deep Neural Networks (DNN), and \textit{angular-penalty} criterion, the state-of-the-art (SOTA) Face Recognition (FR) approaches have performed exceedingly well \cite{wen2021sphereface2,deng2019arcface,wang2018cosface}. 
However, the release of new evaluation benchmarks reveals the performance bias in favor of independent and identical distributed (i.i.d.) data, \ie, train and test instances ought to be drawn from the same distribution \cite{liu2022controllable,kim2022adaface,shi2019probabilistic}. 
Specifically, available FR training sets represent Semi-Constrained (SC) distribution and a considerable performance drop is observed on Un-Constrained (UC) benchmarks \cite{robbins2022effect}, \eg, performance on IJB-S \cite{1} and TinyFace \cite{2} (SC benchmarks), are about 30\% lower than that on LFW \cite{huang2008labeled}.
A straightforward remedy is constructing a large-scale training dataset with sufficient UC and SC samples. However, obtaining a scaleable dataset containing a balanced number of SC and UC samples is infeasible \cite{terhorst2023qmagface}.

\begin{figure}[t]
  \centering
    \includegraphics[width=1.0\linewidth]{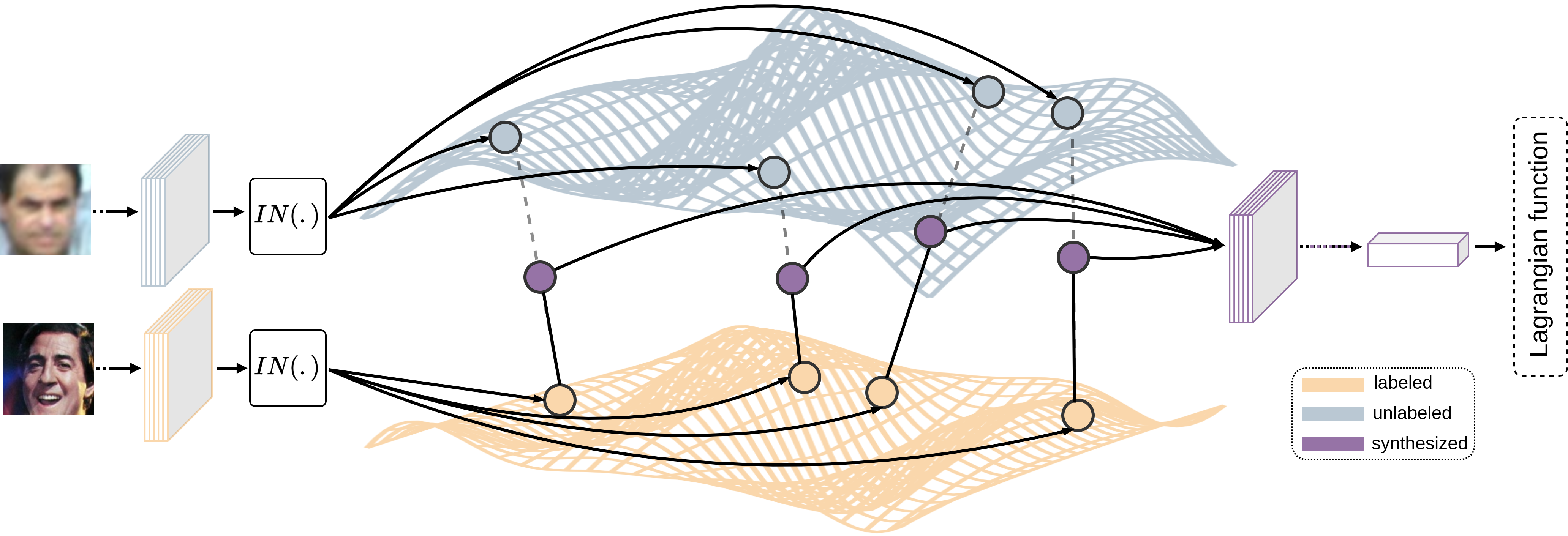}

   \caption{Overall pipeline of the proposed method. We synthesize diverse,
plausible, and novel styles by combining style information from the labeled and unlabeled samples. We move the labeled samples (yellow) toward unlabeled (blue) instances in the style space while ensuring that we do not remove identity-related information from the final embedding, \ie, push away from unrecognizable clusters.
The circles reflect the style information.  }
   \label{onecol}
\end{figure}

As a viable alternative, the mainstream works map UC instances to their SC counterparts
\cite{chen2018fsrnet,yu2018face,ma2020deep,meishvili2020learning,kalarot2020component,ge2020efficient,zangeneh2020low,wang2019improved,singh2019dual}.
These methods can be categorized into: 1) Facial Image Enhancement (FIE) \cite{chen2018fsrnet,yu2018face,ma2020deep,meishvili2020learning,kalarot2020component}, and 2) Pair-wise Common Space Mapping (PCSM) \cite{ge2020efficient,zangeneh2020low,wang2019improved,singh2019dual}.
Although FIE studies improve the visual quality of UC inputs, FIE is inherently ill-posed since multiple SC faces exist for a single UC image \cite{yang2014single,yang2019deep,huang2017beyond}. 
Furthermore, FIE studies are not tailored to FR, which limits their applicability in boosting FR performance \cite{liu2022controllable}. 
PCSM methods try to improve performance by explicitly forcing UC instances to be mapped near the SC samples \cite{ge2020efficient}. These methods require negative and positive pairs of (SC, UC) samples to be available during training, limiting their utility due to the combinatorial explosion in
the number of possible pairs in large-scale setups \cite{ahonen2008recognition,wang2019improved, saas}.

Recent studies have focused on the integration of image augmentation and conventional angular-margin \cite{kim2022adaface,deng2019arcface,shi2021boosting,liu2022controllable}. Despite promising \cite{kim2022adaface}, classical augmentations either do not reach the full potential of UC data or are computationally expensive \cite{robbins2022effect}. Very recently, \cite{shi2021boosting,liu2022controllable} employed unlabeled datasets to generate UC stylized versions of SC samples; here, we dubbed them Image Space Targeted Augmentation (ISTA).
Although innovative, ISTA approaches are fundamentally limited as Generative Adversarial Networks (GANs) are cumbersome to use \cite{yang2022adversarial,li2023rethinking}. Furthermore, training a generative model is not a trivial task and requires relying on strong prior knowledge \cite{antoniou2017data,yang2022adversarial}.

\begin{figure}[t]
  \centering
    \includegraphics[width=1.0\linewidth]{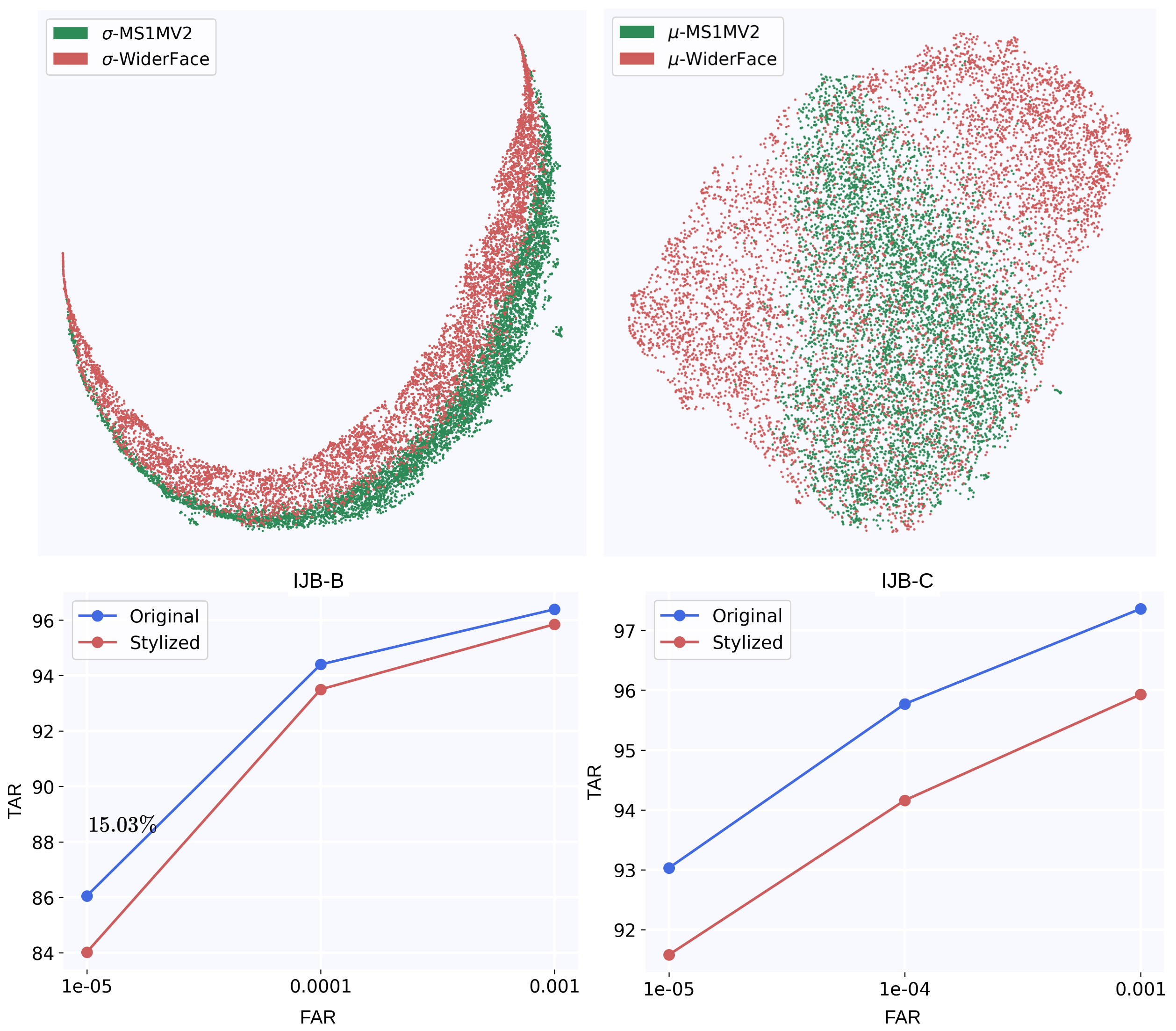}

   \caption{Effect of UC style information on the FR model.
   Top: Illustrates the UMAP visualization of instance-wise $\small{\mu}$ and $\small{\sigma}$ (Equations \ref{mu},and \ref{sigma}) of the output of the 3rd block of the ResNet-50, emphasizing the disparity between the SC (MS1MV2) and UC (WiderFace) datasets.
   Bottom: The performance of the FR model on the original IJB-B/IJB-C and their style-perturbed version. We swapped the IJB-B/IJB-C style information with those from WiderFace. The results show the FR model's susceptibility to the style of the input.  }
   \label{motivationss}
\end{figure}

In this work, we argue that the effect of the input distribution (UC/SC) in the style feature is an essential factor that the model overfits; see Figure \ref{motivationss}. 
To combat this, it is not trivial to: 1) construct more diverse training without employing generative models so it can be integrated into large-scale training and 2) assemble a flexible augmentations module to adjust \textit{w.r.t} the optimization trajectory and produce plausible styles. In this regard, we utilize Domain Generalization (DG) \cite{zhou2021domain} and Adversarial Training (AT) \cite{volpi2018generalizing}, which have shown exceptional progress in out-of-domain (OOD) generalization \cite{zhou2021domain,volpi2018generalizing}. 
However, these techniques are mainly designed for a close-set setup where the label space is consistent across the datasets, while FR is an open-set problem, which results in a multitude of possibilities for how to boost FR with these ideas. 

We introduce a novel method dubbed Targeted Style Augmentation (TSA), which operates within the model's hidden space. Notably, TSA distinguishes itself from contemporary DG/AT methods due to: 1) the open-set nature of FR, where target classes remain disjoint for each dataset, and 2) the introduction of a novel adversarial objective tailored to FR, which augments training diversity while upholding the plausibility of generated style. We present an innovative style augmentation strategy meticulously designed to produce diverse yet plausible styles that deviate from the original SC distribution. This augmentation process is carried out within the model's hidden space, effectively sidestepping the computational overhead and complexities associated with image space manipulation. Moreover, our approach incorporates an adaptive capability for discerning unrecognizable instances, ensuring that the augmentation process aligns seamlessly with the training trajectory.
Our main contributions are three-fold:
\begin{itemize} 
    \item We present a method to manipulate the style information of the features obtained from SC samples while ensuring the validity of generated styles. 
    \item We propose an entropy-based approach to discern the unrecognizable instances during the training. 
    \item We present a targeted augmentation technique with much less computational overhead compared to its GAN-based predecessors, which is tailored for large-scale FR training. 
\end{itemize}

\section{Related Works}
\label{sec:relatedworks}

\subsection{Unconstrained FR}
Most recent works map UC instances to their SC counterparts
\cite{chen2018fsrnet,yu2018face,ma2020deep,meishvili2020learning,kalarot2020component,ge2020efficient,zangeneh2020low,wang2019improved,singh2019dual}, and can be categorized into: 1) Facial Image Enhancement (FIE) \cite{chen2018fsrnet,yu2018face,ma2020deep,meishvili2020learning,kalarot2020component}, and 2) Common Space Mapping (CSM)  \cite{ge2020efficient,zangeneh2020low,wang2019improved,singh2019dual}.
FIE methods have focused on solving the inverse problem of retrieving high-quality (SC) samples from their low-quality (UC) counterparts \cite{yang2014single,yang2019deep,huang2017beyond}. Despite successful \textit{w.r.t} human perception, they fail to increase the FR performance on real UC testing \cite{kim2019progressive,yang2019fsa,liu2022controllable}. 
A different avenue of investigation, CSM, tries to improve the UC samples' representation discriminability by finding joint embedding for (UC, SC) pairs. However, large-scale web-crawled datasets lack the mentioned pair-wise tuplet. Also, such sample-level supervision is unstable and data-inefficient \cite{wen2021sphereface2}. Therefore, CSM application is limited to small-scale controlled datasets \cite{ahonen2008recognition,2,wang2019improved}.

Recently, Image Augmentation (IA) \cite{kim2022adaface,shi2021boosting,liu2022controllable} methods have emerged as an alternative branch of inquiry. Kim \etal \cite{kim2022adaface} use random down-sampling/cropping, which results in a considerable performance improvement on UC benchmarks such as TinyFace. However, the employed augmentation is an oversimplified version of in-the-wild image distortions \cite{robbins2022effect}.
On the other hand, more sophisticated classical augmentation is computationally expensive to be integrated into the FR framework. In \cite{shi2021boosting}, the author proposes to employ GAN as an augmentation module. Followup work \cite{liu2022controllable}, integrate GAN with the quality module to control the level/direction of applied degradation. The major limitation of GAN-based augmentation is that they are cumbersome to deploy during training \cite{yang2022adversarial}.  Additionally, developing a proper generative model heavily relies on prior knowledge, and preserving the identity information during GAN-based augmentation is controversial \cite{wang2019improved}.   

\subsection{Domain Generalization}
The remarkable achievements of deep learning hinge on the premise that the training and test data are derived from an identical distribution \cite{zhong2022adversarial}. When this assumption is violated, considerable performance degradation occurs \cite{robbins2022effect}. DG methods address the cross-domain challenge of learning a generalizable model when multiple source datasets are accessible \cite{zhou2021domain}. Conventional DG methods employ auxiliary losses, such as Maximum Mean Discrepancy or domain adversarial framework \cite{zhou2021domain}. 
Recently, \cite{sankaranarayanan2018generate,zhu2017unpaired,hoffman2018cycada} employ image generation modules to increase the diversity of the training set by mapping the images from one domain to another arbitrary domain while maintaining the content of the original input. 
A new line of study considers the DG as the problem of feature distribution matching \cite{zhou2021domain,kang2022style,zhang2022exact}. Specifically, they augment the cross-distribution feature by utilizing first and second-order statistics. Style perturbation is also utilized in our method as well, but it is being applied in an open-set setup. Furthermore, DG approaches do not deal with degraded imagery to the point that they are not recognizable. We specifically manipulate the augmentation objective to ensure the plausibility of the generated style.

\subsection{Adversarial Training in DG and FR}
Adversarial attacks have brought to light the susceptibility of deep learning models to imperceptible perturbations, as demonstrated by Jia \etal \cite{jia2022adv}. Subsequently, the pioneering work of Goodfellow \etal  \cite{goodfellow2014explaining} proposed Adversarial Training (AT) as a pivotal strategy to mitigate the impact of these perturbations. Recently, the Domain Generalization (DG) incorporated with gradient-based AT has revolutionized DG, as showcased by Volpi \etal \cite{volpi2018generalizing} and Qiao \etal \cite{qiao2020learning}. Within this context, several studies, including Qiao \etal \cite{qiao2020learning,qiao2021uncertainty} and Fan \etal \cite{fan2021adversarially}, have incorporated meta-learning and adaptive normalization into AT to cultivate domain-invariant representations.

Broadly, AT in FR strives to enhance the model's resilience against certain adversarial components in the inputs. Notably, Sharif \etal \cite{sharif2016accessorize} devised real-world adversarial attacks involving printed glasses, while Komkov \etal \cite{komkov2021advhat} explored physical attacks through adversarial hats. Deb \etal \cite{deb2020advfaces} initially leveraged Generative Adversarial Networks (GANs) for synthesizing adversarial samples. Subsequently, Yin \etal \cite{yin2021adv} introduced an adversarial makeup generation framework, while Dong \etal \cite{dong2019efficient,dong2018boosting} harnessed generative models for creating adversarial attributes. Liu \etal \cite{liu2022controllable} harnessed adversarial manipulation of the GAN hidden state to produce challenging training instances for deep FR models. However, it's worth noting that the computational cost associated with generating adversarial examples using GANs renders these GAN-based approaches impractical for large-scale problems, as observed by Yang \etal \cite{yang2022adversarial}. Additionally, the process of training a generative model capable of preserving identity information while transferring from seen to unseen images necessitates a substantial amount of prior knowledge \cite{antoniou2017data}.

\section{Problem Formulation}
\label{sec:method}
 \textbf{Notation.}
Let $\small{\mathcal{D}={(\bx_i,y_i)}_{i=1}^{|\mathcal{D}|}}$ and $\small{\widehat{\mathcal{D}}={({\widehat{\bx}}_i)}_{i=1}^{|\widehat{\mathcal{D}}|}}$ represent the labeled and unlabeled datasets, respectively, where $\small{|.|}$ reflects the cardinality of the set. Here, we decompose backbone $\small{E_{\theta}(.)}$ into two sub-networks $\small{E=E_2 \circ E_1 }$ with trainable parameter $\small{\theta= [\theta_1,\theta_2]}$. $\small{E_1}$ maps the input image into intermediate feature map $\small{\bh=E_1(\bx) \in  \mathbb{R}^{c \times h \times w}}$ where $\small{c}$, $\small{h}$, and $\small{w}$ indicate the channel, height, and width of $\small{\bh}$, while $\small{E_2}$ maps the feature maps $\small{\bh}$ to $\small{d}$-dimensional embedding vector $\small{\bz =E_2(\bh) \in \mathbb{R}^d}$.
Suppose trainable parameter (prototypes) $\small{\bW=[\bw_1,\bw_{2}, \dots,\bw_{c}]}$, refers to the classifier head, that maps $\small{\bz}$ to probability distribution over $c$ classes.
For the convenience of presentation, we omit the bias from the classifier head, and all the representations and prototypes are $\ell_2$-normalized.

\begin{figure*}[t]
  \centering
    \includegraphics[width=1.0\linewidth]{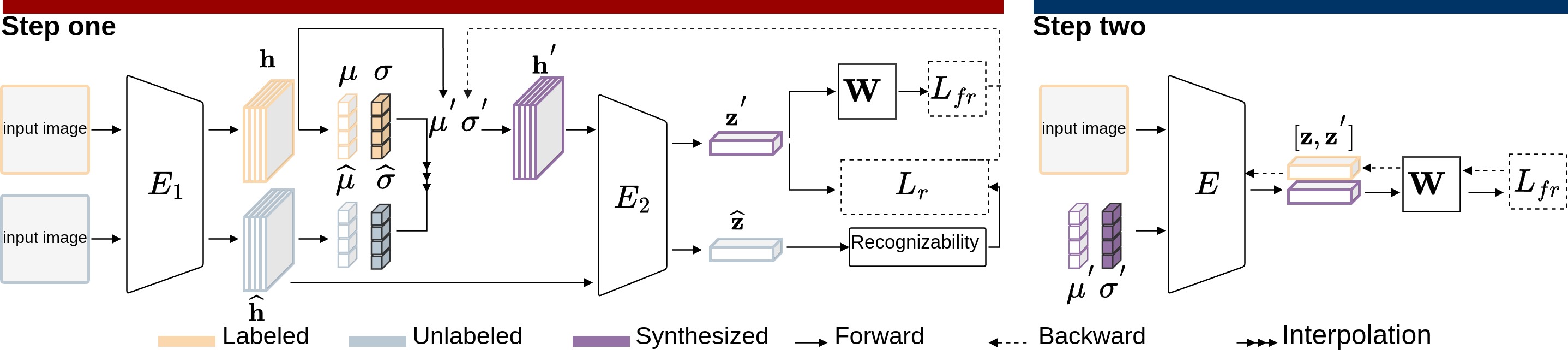}

   \caption{First Stage: the backbone $\small{E = E_2(E_1)}$ is fixed. Two samples from labeled and unlabeled datasets are forwarded to the $\small{E_1}$. Then, a novel style is computed from a combination of labeled and unlabeled style information and finally reconstructs the $\small{\bh^{'}}$ with the novel style. Then, the FR loss function, $\small{L_{fr}}$, and the recognizability loss is computed, and their gradient are used to optimize the interpolation coefficients. Second Stage: The labeled data and synthesized feature map, $\small{\bh^{'}}$, are used to train the FR model $E$.}
   \label{main_diagram}
\end{figure*}

\subsection{Preliminary}\label{PRE}
\textbf{Style.}\label{SE}
The pioneering work of \cite{ulyanov2016instance} discovers that instance-specific channel-wise mean and standard deviation of intermediate feature map from a CNN convey the style information of an input image, dubbed as $\small{\bmu=[\mu_1,\mu_2, \dots,\mu_c]}$ and $\small{\bsigma=[\sigma_1,\sigma_2, \dots,\sigma_c]}$, respectively. Accordingly, given the $\small{\bh}$, the style information can be obtained by:
\begin{equation}\label{mu}
 \small
 \begin{aligned}
  \mu_{k} = \frac{1}{hw}\sum_{i=1}^h\sum_{j=1}^w {\bh_{k,i,j}},
\end{aligned}
\end{equation}
\begin{equation}\label{sigma}
 \small
 \begin{aligned}
  \sigma_{k} = \sqrt{\frac{1}{hw}\sum_{i=1}^h\sum_{j=1}^w {(\bh_{k,i,j}-\mu_k})^2},
\end{aligned}
\end{equation}
where subscript $\small{k}$ refers to the channel along which the computation is applied. With the application on neural style transfer, AdaIN \cite{huang2017arbitrary} replaces the style of the original feature map $\small{\bh}$ with target style $\small{(\widehat{\bmu},\widehat{\bsigma})}$:
\begin{equation}\label{adain}
 \small
 \begin{aligned}
  {\bh}^{'} = \widehat{\bsigma} \frac{\bh-\bmu}{\bsigma}+\widehat{\bmu}.
\end{aligned}
\end{equation}

Our method draws inspiration from the AdaIN application in generative models. However, rather than generating an image with a novel style, we leverage adversarial perturbation in the model's hidden style space to implicitly increase the hardness of the training samples and force SC instances to illustrate the characteristics of UC images.

\vspace{5pt}
\noindent\textbf{Adversarial training.}\label{GBAA}
Given the dataset $\small{\mathcal{D}}$, adversarial training \cite{goodfellow2014explaining} can be formulated as:
\begin{equation}\label{pgd}
 \small
 \begin{aligned}
  \mathop{{\min}}_{{\btheta,{\bW}}}{\mathbb{E}_{(\bx,y)\in \mathcal{D}}{\left[L(E(\bx),\bW, y) + \mathop{{\max}}_{{||\bdelta||_p\leq \epsilon}}\: L(E(\bx+\bdelta),\bW, y)\right]}},
\end{aligned}
\end{equation}
where $\small{\bdelta}$ is the crafted perturbation constrained within a $\ell_p$-norm ball centered at  $\small{\bx}$ with a radius $\epsilon$, and $\small{L_{adv}}$ is the adversarial cost function \cite{goodfellow2014explaining}. The perturbation $\small{\bdelta}$ can be crafted by multi-step Projected Gradient Descent (PGD) \cite{madry2017towards} on the linearized cost function:  
\begin{equation}\label{pgd_training}
 \small
 \begin{aligned}
  \bdelta_{t+1}= \Pi_{||\bdelta||_{p}\leq \epsilon}{\left[\bdelta_t + \alpha \sgn(\nabla_{\bdelta_t}L(E(\bx+\bdelta_t),\bW,y)\right]},
\end{aligned}
\end{equation}
where $\small{\sgn}$ denotes the sign function and $\small{\alpha}$ is the inner maximization step size. Performing the inner-maximization of Equation \ref{pgd} using Equation \ref{pgd_training} results in $\bdelta$ that alters $\bx$ to increase $L$, \ie, producing hard training instance. Then, the outer minimization of Equation \ref{pgd} trains $E$ using this hard instance to encourage the robustness to adversarial components presenting in $\bx+\bdelta$.

\subsection{Recognizability}\label{IPC}
Conventionally, $\small{L}$ in Equation \ref{pgd} is the task cost function, and adversarial manipulation is performed in the image space by directly changing the pixels' intensity \cite{goodfellow2014explaining,shin2022teaching,  yang2022adversarial, li2023rethinking, antoniou2017data}. Instead of altering the pixels' intensity, we manipulate the intermediate features obtained from SC inputs toward representing UC characteristics. To this end, we interpolate between the style of unlabeled (UC) and labeled (SC) samples. However, unlabeled face datasets naturally comprise a large amount of detectable but unrecognizable faces \cite{deng2023harnessing}. Thus, adversarial interpolation using gradient ascent on the training objective, may result in unrecognizable styles, which has detrimental effects on model training. We propose a metric to constrain the recognizability of the features resulting from the synthesized style.

Recent studies \cite{robbins2022effect,deng2023harnessing,chai2023recognizability} reveal that unrecognizable instances form a cluster well-separated from other identities, dubbed UR cluster. 
Inspired by this finding, we measure the recognizability as the function of distance to the UR cluster:
\begin{equation}\label{constraint}
 \small
 \begin{aligned}
 L_r = \frac{1}{\epsilon+1-\left<{\bphi},{E_2(\widehat{\bh})}\right>},
\end{aligned}
\end{equation}
where $\left<.,.\right>$ represents the inner product, and $\small{\bphi}$ denotes the centroid of UR cluster. The first step for computing $\small{L_r}$ is to compute the $\small{\bphi}$, which requires to distinguish UR instances. \cite{deng2023harnessing,chai2023recognizability} detect UR samples and precompute $\small{\bphi}$ in an offline manner before the training. We argue that since the encoder $\small{{E}}$ evolves progressively, synchronizing $\small{\bphi}$ with the training trajectory is essential.
Thus, we aim to distinguish unrecognizable samples through the information entropy of their representation as the training progresses \cite{thomas1991elements,pal1991entropy,dan2023transface,shannon2001mathematical}.

Specifically, we aim to measure the recognizability of a feature vector $\small{\bz_i}$ by its information entropy $\small{H{(\bz_i)}}$. 
Yet, feature embedding of a deep network follows a complex and unknown distribution, and it is infeasible to explicitly compute the $\small{H{(\bz_i)}}$ \cite{dan2023homda,guo2021sample}.
As a viable alternative, it has been proven that under the Maximum Entropy Principle, the entropy of an arbitrary distribution is upper bounded by a Gaussian with the same mean and variance \cite{sun2021mae,kullback1997information,jaynes1957information}. In particular, if $\small{z}$ is sampled from a Gaussian with mean $\small{m}$ and variance $\small{v^2}$, $\small{ z\sim\mathcal{N}(m, v^2)}$, the upper bound to differential entropy of $\small{z}$ can be defined as:
\begin{equation}\label{entropy}
\small
\begin{aligned}
H(z)&= -\int{p(z)\log{p(z)}dx} \\
   &= -\mathbb{E}[\mathcal{N}(m,v^2)]\\
   &=  -\mathbb{E}[\log{[(2\pi v^2)^{\frac{-1}{2}}e^{(\frac{-1}{2v^2}(z-m)})]}]\\
   &= \frac{1}{2}\mathbb{E}[\log{(2\pi v^2)}] +\frac{1}{2v^2}\mathbb{E}[(z-m)]\\ 
   &= \frac{1}{2}\log{2\pi v^2}+\frac{1}{2}, 
\end{aligned}
\end{equation}
which is solely a function of $v$. Therefore, we estimate the upper bound of the entropy (information) of representation $\small{\bz_i}$ using its variance. 

\begin{figure}[t]
  \centering
    \includegraphics[width=1\linewidth]{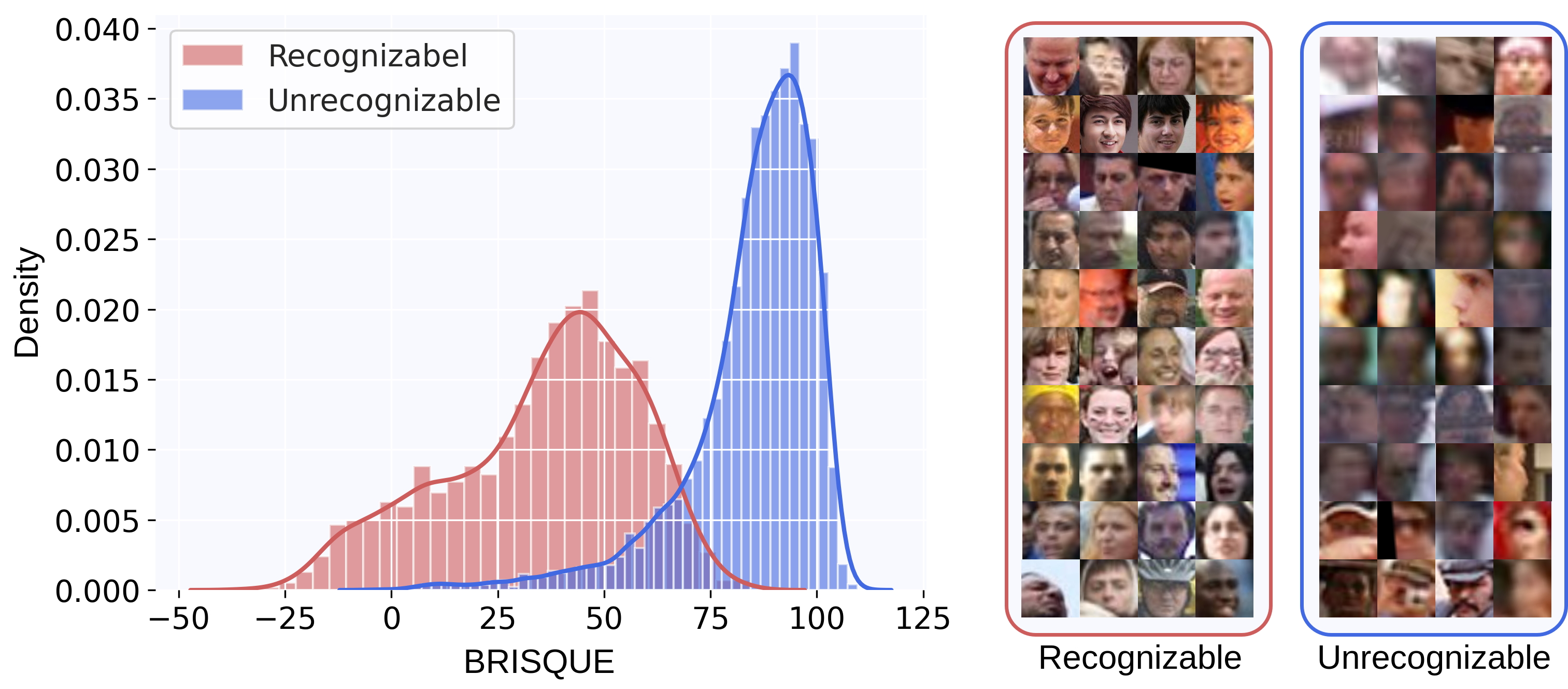}

   \caption{
   Left: Histogram of BRISQUE scores for both unrecognizable and recognizable instances from the WiderFace dataset drawn from the final training epoch. Note that a lower BRISQUE score is indicative of higher image quality. This histogram demonstrates that the presented approach to distinguish unrecognizable instances effectively characterizes facial quality. Right:  Demonstrating randomly sampled unrecognizable and recognizable images from WiderFace.  While it is important to acknowledge that some recognizable instances may be mistakenly classified within the UR cluster, our primary interest lies in the center of this cluster, which is representative of the majority.}
   \label{ur_cluster}
\end{figure}

For a batch of unlabeled face images, we first estimate the entropy of each sample using Equation \ref{entropy}. Then, we compute the exponential moving average over embedding samples with the top-$\small{k}$ lowest entropy \cite{he2020momentum,lucas1990exponentially}, $\small{\overline{\bz}_{topk}}$:
\begin{equation}\label{mu_t}
\small
 \begin{aligned}
 &\bphi^{t+1} = \alpha \bphi^{t} + (1-\alpha) \overline{\bz}_{topk},
\end{aligned}
\end{equation}
where $\small{\alpha} \in [0,1]$ is the momentum coefficient. Figure \ref{ur_cluster} experimentally shows the unrecognizable samples detected by our simple yet effective entropy-based selection approach. Besides, we plot the image quality score (BRISQUE) to compare the quality of unrecognizable and recognizable instances, illustrating the efficiency of the proposed approach in distinguishing samples based on their recognizability.

\subsection{Targeted Style Adversary}\label{SA}
Here, we present the proposed Targeted Style Adversarial training pipeline.
Specifically, given the style information from labeled $\small{(\bmu,\bsigma)}$ and unlabeled $\small{(\widehat{\bmu},\widehat{\bsigma})}$, novel style can be computed by their convex combination:
\begin{equation}\label{sigma_new}
 \small
 \begin{aligned}
  {\bsigma}^{'} = \lambda_1 \bsigma + (1-\lambda_1) \widehat{\bsigma},
\end{aligned}
\end{equation}
\begin{equation}\label{mu_new}
 \small
 \begin{aligned}
  {\bmu}^{'} = \lambda_2 \bmu + (1-\lambda_2) \widehat{\bmu},
\end{aligned}
\end{equation}    
where $\small{\lambda_1}$ and $\small{\lambda_2}$ define the amount each labeled style should move toward that of unlabeled. Then, optimal $\small{\lambda_1}$ and $\small{\lambda_2}$ are obtain using PGD with $L$ defined as:
\begin{equation}\label{advobj}
 \small
 \begin{aligned}
 L = L_{fr} - \beta \: L_r,
\end{aligned}
\end{equation}
where $L_{fr}$ is an arbitrary FR objective function, \eg, ArcFace \cite{deng2019arcface}. The objective in Equation \ref{advobj} has a similar form as the original cost function for the gradient-based adversarial attack in that they both strive to find the optimal augmented version of the clean input. Additionally, here the augmentation function needs to minimize the constraint, in addition to maximizing $\small{L_{fr}}$. The augmentation that is obtained in this manner avoids collapsing the inherent identity-related information because the constraint will explode when the transformed representation drifts toward the $\small{\bphi}$. In essence, Equation \ref{advobj} requires the augmentation function to transform style so that they are hard for the FR model in the range where valid for FR. Subsequent to obtaining the desired synthetic style information, the FR model is training using original and synthesized styles, as illustrated in Figure \ref{main_diagram} and Algorithm \ref{alg1}.

\begin{algorithm}[t]

\small
\caption{TSA}\label{alg1}
Initialize $E$, $\bW$, $\small{\alpha>0}$, $\small{\lambda_1}$, $\small{\lambda_2}$, $\small{\beta>0}$, $\small{k>0}$, and total number of training iterations $\small{t_1>0}$. 

\For{$\small{t=0}$ ... $\small{t_1\!-\!1 }$}{
$\mathcal{B} \: \leftarrow$ Sample batch of labeled images

$\widehat{\mathcal{B}} \: \leftarrow$ Sample batch of unlabeled images

$\small{\bh=E_1(\mathcal{B})}$

$\small{\widehat{\bh}=E_1(\mathcal{\overline{B}})}$

$\small{\widehat{\bz}=E_2(\widehat{h})}$

$\small{\bmu, \bsigma \leftarrow }$ style information from $\bh$

$\small{\widehat{\bmu}, \widehat{\bsigma}  \leftarrow }$ style information from $\widehat{\bh}$

$\bphi \leftarrow $ distinguish un/recognizable unlabeled instances 

\For{$\small{i=0}$ ... $\small{k\!-\!1 }$}{
${\bsigma^{'} \: \leftarrow \: \lambda_1 \bsigma + (1-\lambda_1) \widehat{\bsigma}}$

$\small{\bmu^{'} \: \leftarrow \: \lambda_2 \bmu + (1-\lambda_2) \overline{\bmu}}$

$\small{\bh^{'} = \bsigma^{'} \frac{\bh-\bmu}{\bsigma}+\bmu^{'}}$

$\small{\bz^{'} = E_2(\bh^{'})}$

Compute $\small{L}$

Update $\small{\lambda_1}$, and $\small{\lambda_2}$ using gradient ascent on Equation \ref{advobj}

}

$\small{\bsigma^{'} \: \leftarrow \: \lambda_1 \bsigma + (1-\lambda_1) \widehat{\bsigma}}$

$\small{\bmu^{'} \: \leftarrow \: \lambda_2 \bmu + (1-\lambda_2) \widehat{\bmu}}$

$\small{\bh^{'} = \bsigma^{'} \frac{\bh-\bmu}{\sigma}+\bmu^{'}}$

$\small{\bz^{'} = E_2(\bh^{'})}$

$\small{\bz = E_2(\bh)}$

Compute $\small{L_{fr}}$

Update $E$ using gradient descent on $\small{L_{fr}}$

}

\end{algorithm}

\section{Experiments}
\label{sec:experiments}

\begin{table}[]
\addtolength{\tabcolsep}{-3.0pt} 
\small

\begin{center}
\begin{tabular}{lcccc}
\toprule
Method         & Dataset & Backbone   & ACC@1 & ACC@5 \\ \hline
CurricularFace & MS1MV2  & ResNet-100 & 63.68 & 67.65 \\
ArcFace  $\star$      & MS1MV2  & ResNet-50  & 62.7  & 65.85 \\
CFSM   $\star$     & MS1MV2  & ResNet-50   & 63.21 & 67.23 \\
TSA           & MS1MV2  & ResNet-50   & \textbf{64.40} & \textbf{68.53} \\ \midrule 
CFSM        & WebFace12M  & ResNet-100   & 73.87 & 76.77 \\
TSA           & WebFace12M  & ResNet-100   & \textbf{74.02} & \textbf{77.57} \\
\bottomrule
\end{tabular}
\vspace{5pt}
\caption{\small{Identification performance on TinyFace compared to the SOTA methods. ' $\star$ ' reflects a re-execution of the official code provided by the authors with the optimal configurations as stated in the original paper.}}\label{TinyFace} 
\end{center}
\end{table}


\subsection{Datasets} 
We utilize cleaned version MS-Celeb-1M \cite{guo2016ms} as our labeled training dataset. The original MS-Celeb-1M includes a significant amount of labeling noise. Pioneering work of \cite{deng2019arcface} provides a cleaned version of this dataset with almost 4M images from 85K identities. As per the conventional FR framework, all used datasets in our work are aligned and transformed to $\small{112 \times 112}$ pixels. For unlabeled datasets, we employed WiderFace \cite{yang2016wider}, which is known to be one of the most challenging benchmarks in face detection studies due to diversity in face scale and nuances. Thus, WiderFace works for our purpose since it contains both recognizable and unrecognizable faces. Following \cite{shi2021boosting,liu2022controllable} and for a fair comparison, we utilize 70k face images from WiderFace.
We report performance on  TinyFace \cite{cheng2019low}, IJB-B \cite{whitelam2017iarpa}, IJB-C \cite{maze2018iarpa}, IJB-S \cite{kalka2018ijb}, and SCFace \cite{grgic2011scface} datasets to evaluate the proposed approach. 

\begin{table}[]
\addtolength{\tabcolsep}{-4pt} 
\small
\begin{center}
\begin{tabular}{lcccccc}
\toprule
\multicolumn{1}{c}{\multirow{2}{*}{Method}} & \multicolumn{1}{c}{\multirow{2}{*}{Dataset}} & \multicolumn{1}{c}{\multirow{2}{*}{Backbone}} & \multicolumn{4}{c}{TAR@FAR}                                                                        \\\cline{4-7} 
\multicolumn{1}{c}{}                        & \multicolumn{1}{c}{}                         & \multicolumn{1}{c}{}                          & \multicolumn{1}{c}{1e-6} & \multicolumn{1}{c}{1e-5} & \multicolumn{1}{c}{1e-4} & \multicolumn{1}{c}{1e-3} \\ \hline
\multicolumn{1}{l}{VGGFace2}                & \multicolumn{1}{c}{VGGFace2}                 & \multicolumn{1}{c}{ResNet-50}                 & \multicolumn{1}{c}{-}    & \multicolumn{1}{c}{70.5} & \multicolumn{1}{c}{83.1} & \multicolumn{1}{c}{90.8} \\
\multicolumn{1}{l}{Comparator}              & \multicolumn{1}{c}{VGGFace2}                 & \multicolumn{1}{c}{ResNet-50}                 & \multicolumn{1}{c}{-}    & \multicolumn{1}{c}{-}    & \multicolumn{1}{c}{84.9} & \multicolumn{1}{c}{93.7} \\
ArcFace  & MS1MV2  & ResNet-50  & 40.77  & 84.28    & 91.66     & 94.81     \\
AFRN    & VGGFace2* & Resnet-100 & -      & 77.1    & 88.5       & 94.9                     \\
MagFace & MS1MV2    & ResNet-50  & -      & 83.87   & 91.47      & 94.67                    \\
Shi et al. & MS1MV2* & ResNet-50 & 43.38  & 88.19   & 92.87     & 95.86                    \\

ArcFace  $\star$  & MS1MV2& ResNet-50 & 40.03 & 86.05 & 93.40   & 95.39                        \\

CFSM  $\star$ & MS1MV2 & ResNet-50     & 40.13& \textbf{90.72} & 94.20   & 96.11                        \\
 TSA & MS1MV2  & ResNet-50     & \textbf{46.47}& 87.22 & \textbf{94.33}   & \textbf{96.32}                        \\ \bottomrule
\end{tabular}
\vspace{5pt}
\caption{\small{Verification on IJB-B compared to the SOTA methods.' * ' signifies specific data subsets chosen by the authors. ' $\star$ ' reflects a re-execution of the official code provided by the authors with the optimal configurations as stated in the original paper.}}\label{verijbb} 
\end{center}
\end{table}

 {\textbf{TinyFace}} \cite{cheng2019low} is a low-quality FR evaluation dataset comprising 5,139 labeled identities with 169,403 images. The images are designed for 1:$N$ recognition tests and have an average size of 20$\times$16 pixels. The images in TinyFace were collected from public web data and captured faces under various uncontrolled conditions, including different poses, illumination, occlusion, and backgrounds.

 \textbf{IJB-B and IJB-C.} IJB-B \cite{whitelam2017iarpa} contains around 21.8K images (11.8K faces and 10K non-face images) and 7k videos (55K frames). A total of 1,845 identities are presented in this dataset. Our experimental protocols follow the standard 1:1 verification, which contains 10,270 positive and 8M negative matches. There are 12,115 templates in the protocol, each of which consists of multiple images or frames. Consequently, a template-based matching process is used. Specifically, we average over the instances in a template to obtain the global feature vector for each template. IJB-C \cite{maze2018iarpa} is the extended version of IJB-B, including 31.3K images and 117.5K frames from 3,531 identities. The testing protocol of IJB-C is similar to IJB-B.

 \textbf{IJB-S} \cite{kalka2018ijb} is one of the most challenging FR benchmarks with samples from surveillance videos. It consists of 350 surveillance videos spanning 30 hours in total, from 202 identities, with an average of 12 videos per subject. Also, there is 7 high-quality photo for each subject with different poses. There are three keywords that define the evaluation metrics in this dataset:
\begin{itemize}
    \item \textbf{Surveillance:} denotes  the surveillance video.
    \item \textbf{Single:} denotes high quality enrollment.
    \item \textbf{Booking:} denotes multiple enrollment images taken from different viewpoints.
\end{itemize}
Evaluation protocols for this data are: 1) \textit{Surveillance-to-Single}, 2) \textit{Surveillance-to-Booking}, and 3) \textit{Surveillance-to-Surveillance}. The first notation indicates the source of gallery inputs, and the second refers to the probe source.

\textbf{SCFace} \cite{grgic2011scface} is a challenging cross-resolution FR evaluation benchmark. It contains mugshots and images captured by surveillance cameras.
The images were taken from 130 subjects in an uncontrolled indoor environment using five video surveillance cameras at three different distances $\{4.2, 2.6, 1.0\}$ (meter); five images at each distance. Also, one frontal mugshot image for each subject is obtained using a digital camera.


\begin{table}[]
\addtolength{\tabcolsep}{-4pt} 
\small

\begin{center}
\resizebox{1\linewidth}{!}{
\begin{tabular}{lcccccc}
\toprule
\multicolumn{1}{l}{\multirow{2}{*}{Method}} & \multicolumn{1}{c}{\multirow{2}{*}{Dataset}} & \multicolumn{1}{c}{\multirow{2}{*}{Backbone}} & \multicolumn{4}{c}{TAR@FAR}                                                                          \\ \cline{4-7} 
\multicolumn{1}{c}{}                        & \multicolumn{1}{c}{}                         & \multicolumn{1}{c}{}                          & \multicolumn{1}{c}{1e-7} & \multicolumn{1}{c}{1e-6} & \multicolumn{1}{c}{1e-5}  & \multicolumn{1}{c}{1e-4}  \\ \hline
\multicolumn{1}{l}{VGGFace2}                & \multicolumn{1}{c}{VGGFace2}                 & \multicolumn{1}{c}{ResNet-50}                 & \multicolumn{1}{c}{-}    & \multicolumn{1}{c}{-}    & \multicolumn{1}{c}{76.8}  & \multicolumn{1}{c}{86.2}  \\
\multicolumn{1}{l}{PFE}                     & \multicolumn{1}{c}{MS1MV*}                   & \multicolumn{1}{c}{ResNet-64}                 & \multicolumn{1}{c}{-}    & \multicolumn{1}{c}{-}    & \multicolumn{1}{c}{89.64} & \multicolumn{1}{c}{93.25} \\
ArcFace                                     & MS1MV2                                       & ResNet-50                                     & 67.40                    & 80.52                    & 88.36                     & 92.52                     \\
AFRN                                        & VGGFace2*                                    & Resnet-100                                    & -                        & -                        & 88.3                      & 93.0                      \\
DUL                                         & MS1MV*                                       & ResNet-64                                     & -                        & -                        & 90.23                     & 94.2                      \\
MagFace                                     & MS1MV2                                       & ResNet-50                                     & -                        & 81.69                    & 88.95                     & 93.34                     \\
Shi et al.                                  & MS1MV2*                                      & ResNet-50                                     & 77.39                    & 87.92                    & 91.86                     & 94.66                     \\

ArcFace  $\star$                                     & MS1MV2                                       & ResNet-50                                     & 76.58                        & 84.38                        & 93.03                         & 95.77                         \\

CSFM    $\star$                                    & MS1MV2                                      & ResNet-50                                     & 77.56                        & \textbf{89.03}                   & \textbf{93.94}                     & 95.77                     \\
TSA                                       & MS1MV2                                       & ResNet-50                                     & \textbf{78.11}                        & 88.53                       & 93.36                         & \textbf{95.79}                         \\ 
\bottomrule
\end{tabular}}
\vspace{5pt}
\caption{\small{Verification on IJB-C compared to the SOTA methods.' * ' signifies specific data subsets chosen by the authors. ' $\star$ ' reflects a re-execution of the official code provided by the authors with the optimal configurations as stated in the original paper.}}\label{verijbc}
\end{center}
\end{table}

\subsection{Implementation Details}
We adopt a modified version of ResNet \cite{deng2019arcface} for the backbone. The model is trained for 24 epochs with ArcFace loss. The optimizer is SGD, with the learning rate starting from 0.1, which is decreased by a factor of 10 at epochs \{10, 16, 22\}. The optimizer weight-decay is set to 0.0001, and the momentum is 0.9. During training, the mini-batch size on each GPU is 512, and the model is trained using four RTX 6000. Following \cite{he2020momentum}, $\small{\alpha}$ in Equation \ref{mu_t} is 0.99.
Given a pair of images, the cosine distance between the representations is the metric during inference.

\begin{figure}[t]
  \centering
    \includegraphics[width=1.0\linewidth]{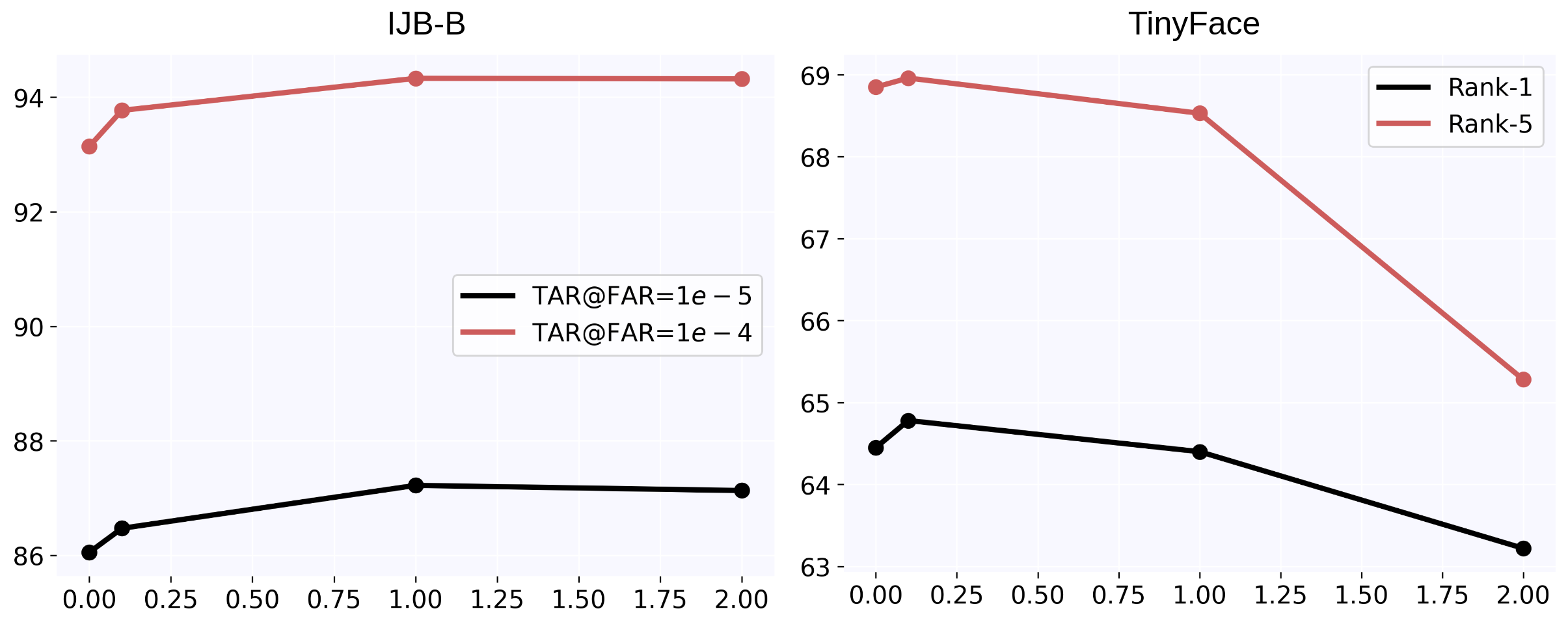}

   \caption{An ablation study on constraining the adversarial objective function. Lowering the constraint, $\small{\beta < 1.0}$, slightly improves TinyFace but degrades IJB-B performance, while a large $\small{\beta}$ benefits neither IJB-B nor TinyFace.}
   \label{betaabb}
\end{figure}

\subsection{Comparison with SOTA Methods}
For a fair comparison, we employed the official code released by ArcFace, and CFSM and utilized the optimal hyper-parameters recommended in their original paper to reproduce their results on MS1MV2. 
Table \ref{TinyFace}, compares the proposed method with others in TinyFace dataset. The presented method outperforms its competitors across evaluation metrics, backbones, and training sets. Specifically, our method surpasses CSFM by 1.19\% and 1.29\% accuracy on rank-1 and rank-5 identification, respectively. 
These improvement emphasise the generalizablity of our framework and that our proposal is not restricted to certain training set or backbone.

Furthermore, Table \ref{verijbb} and Table \ref{verijbc} present the verification performance on IJB-B and IJB-C datasets, illustrating that the proposed approach achieves new SOTA performance.
These consistent improvements on IJB-B and IJB-C showcase the generalizability of our approach on both SC and UC evaluation samples, since these datasets consists SC and UC instances. 
To evaluate the proposed method on cross-resolution scenario, we report evaluation on SCFace. As shown in Table \ref{idscface}, our approach can effectively map both high and low-quality samples of a given identity close to each other and obtain new SOTA performance in this evaluation benchmark. Finally, to evaluate the presented approach in a more challenging scenario, Table \ref{verijbs} shows the performances of IJB-S demonstrating on-par performance with CFSM. It is worth mentioning that we reduce the computational burden of CFSM by avoiding image space manipulation.


\begin{table}[]
\addtolength{\tabcolsep}{-1pt} 
\small

\begin{center}
\resizebox{1\linewidth}{!}{
\begin{tabular}{lccccc}
\hline
\multirow{2}{*}{Method} & \multirow{2}{*}{Venue} & \multicolumn{4}{c}{Rank-1 IR(\%)} \\ \cline{3-6} 
                        &                        & 4.2m   & 2.6m   & 1.0m   & Avg    \\ \hline
T-C                     & IVC20                  & 70.20  & 94.70  & 98.10  & 87.33  \\
TCN                     & ICASSP10               & 74.60  & 94.90  & 98.60  & 89.37  \\
DRVNet                  & TPAMI21                & 76.80  & 92.80  & 97.50  & 89.03  \\
FAN                     & ACCV19                 & 77.50  & 95.00  & 98.30  & 90.30  \\
RAN                     & ECCV20                 & 81.30  & 97.80  & 98.80  & 92.62  \\
MIND-Net                & SPL21                  & 81.75  & 98.00  & 99.25  & 93.00  \\
DDL                     & ECCV20                 & 86.80  & 98.30  & 98.30  & 94.40  \\
RIFR                    & T-BIOM20               & 88.30  & 98.30  & 98.60  & 95.00  \\
RPCL                    & NN22                   & 90.40  & 98.00  & 98.00  & 95.46  \\
IDEA-Net                & TIFS22                 & 90.76  & 98.50  & 99.25  & 96.17  \\
NPT                     & TPAMI22                & 85.69  & 99.08  & 99.08  & 96.61  \\
DSN                     & APSIPA21               & 93.00  & 98.50  & 98.50  & 96.70  \\
AdaFace                 & CVPR21                 & 95.38  & 98.46  & 99.84  & 97.89  \\
Chai et al.             & CVPR23                 & 97.07  & 99.23  & \textbf{99.80}  & 98.70  \\
TSA               &     & \textbf{97.91}                      & \textbf{99.47}      & 99.71      & \textbf{99.03}     \\ \hline
\end{tabular}}
\vspace{5pt}
\caption{\small{Cross-resolution evaluation  on SCFace.}}\label{idscface}
\end{center}
\end{table}



\begin{table*}[]
\addtolength{\tabcolsep}{-3pt} 
\small

\begin{center}
\resizebox{1\linewidth}{!}{
\begin{tabular}{lllllllllllllll}
\toprule
\multirow{2}{*}{Method} & \multirow{2}{*}{Dataset} & \multirow{2}{*}{Backbone} & \multicolumn{4}{c}{V2S}       & \multicolumn{4}{c}{V2B}       & \multicolumn{4}{c}{V2V}      \\
                        &                          &                           & Rank1 & Rank5 & 1     & 10    & Rank1 & Rank5 & 1     & 10    & Rank1 & Rank5 & 1    & 10    \\ \hline
\multicolumn{1}{l}{MARN}                   & MS1MV*                   & ResNet-64                 & 58.14 & 64.11 & 21.47 & -     & 59.26 & 65.93 & 32.07 & -     & 22.25 & 34.16 & 0.19 & -     \\
\multicolumn{1}{l}{PFE}                      & MS1MV*                   & ResNet-64                 & 50.16 & 58.33 & 31.88 & 35.33 & 53.60 & 61.75 & 35.99 & 39.82 & 9.20  & 20.82 & 0.84 & 2.83  \\
\multicolumn{1}{l}{ArcFace}                 & MS1MV2                   & ResNet-50                 & 50.39 & 60.42 & 32.39 & 42.99 & 52.25 & 61.19 & 34.87 & 43.50 & -     & -     & -    & -     \\
\multicolumn{1}{l}{Shi et al.}               & MS1MV2*                  & ResNet-50                 & 59.29 & 66.91 & 39.92 & 50.49 & 60.58 & 67.70 & 32.39 & 44.32 & 17.35 & 28.34 & 1.16 & 5.37  \\
\multicolumn{1}{l}{CFSM}                     & MS1MV2*                  & ResNet-50                 & 63.86 & 69.95 & 47.86 & 56.44 & 65.95 & 71.16 & 47.28 & 57.24 & 21.38 & 35.11 & 2.96 & 7.41  \\ \hline \addlinespace
\multicolumn{1}{l}{TSA}                     & MS1MV2                   & ResNet-50                 & \textbf{65.02}     & \textbf{70.25}     & \textbf{48.11}     & \textbf{56.36}    & \textbf{67.85}     & \textbf{71.74}    & \textbf{49.00}     & \textbf{57.65}     & \textbf{22.01}     & \textbf{35.54}     & \textbf{3.89}    & \textbf{8.45}     \\ 
\bottomrule
\end{tabular}}
\vspace{5pt}
\caption{\small{Comparison with SOTA methods on the IJB-S benchmark.
‘*’ denotes that we re-run the code due to the unavailability of the mentioned method trained on the specific data in the official GitHub repository.}}\label{verijbs}
\end{center}
\end{table*}


\begin{table}[]
\addtolength{\tabcolsep}{-2pt} 
\small

\begin{center}
\resizebox{1\linewidth}{!}{
\begin{tabular}{lcccccc}
\toprule
\multirow{2}{*}{Method} & \multirow{2}{*}{TSA} & \multirow{2}{*}{Dataset} & \multicolumn{2}{c}{IJB-B} & \multicolumn{2}{c}{TinyFace} \\ \cline{4-7} 
                             &                       &                          & 1e-5        & 1e-4        & Rank1         & Rank5        \\ \hline

\multirow{2}{*}{CosFace}     & -                     & MS1MV2                    &85.28           & 93.80           & 61.50             &66.33            \\
                             & \checkmark                 & MS1MV2                    & 88.82           & 94.74           & 63.25             & 67.86            \\ \hline
\multirow{2}{*}{ArcFace}     & -                     & MS1MV2                    & 86.05          & 93.40           & 62.70             & 65.85            \\
                             & \checkmark                 & MS1MV2                    & \textbf{87.22}           & \textbf{94.33}           & \textbf{64.40}             & \textbf{68.53}            \\ \hline
\multirow{2}{*}{AdaFace}     & -                     & MS1MV2                    & 86.32          & 94.14          & 65.10            & 68.84            \\
                             & \checkmark                 & MS1MV2                    & \textbf{87.47}           & \textbf{94.40}          & \textbf{65.95}             & \textbf{68.96}            \\ \bottomrule
\end{tabular}}
\vspace{5pt}
\caption{\small{Orthogonal improvements to SOTA angular margin loss functions by using the proposed method. Reporting TAR@FAR for IJB-B and Identification performance for TinyFace.}}\label{orthoimprove} 
\end{center}
\end{table}

\subsection{Orthogonal Improvement to Angular Margins}
The proposed method works by adjusting sample difficulty for the model training. Therefore, we want to investigate the result of different angular margin loss functions on its effectiveness. Specifically, we adjusted the training code of the following SOTA methods to add our augmentation policy: CosFace ($\small{margin=0.35}$), ArcFace ($\small{margin=0.5}$), and AdaFace ($\small{margin=0.4}$). As shown in Table \ref{orthoimprove}, the proposed method improves the performance across loss functions, indicating the orthogonality of the proposed method to existing angular penalty losses.

\subsection{Effect of Constraint}
To validate the role of constraint in producing valid training instances for the FR model, we train models with varying $\small{\beta}$. Fig. \ref{betaabb} shows the performance of different models. Accordingly, lowering the constraint, i.e., $\small{\beta < 1.0}$, brings marginal improvement in TinyFace; however, it causes performance degradation on IJB-B, showcasing that the model generalization across samples with better quality is reduced. On the other hand, too much restriction, large $\small{\beta}$, produces easy samples for the training and does not significantly benefit performance in either IJB-B or TinyFace.

\subsection{Blind VS. Targeted Adversary}
To better understand the efficacy of targeted augmentation in improving model generalization, we conduct experiments where the procedure of producing an adversarial style does not involve utilizing unlabeled samples, and the style is blindly produced with arbitrary direction. Table \ref{orthoimprove2} compares the performance of the proposed method and non-targeted attack, reflecting the significance of employing unlabeled samples during the training. Note that without employing unlabeled samples, the lagrangian constraint is not accessible, and we don't have access to the unrecognizable samples. We argue that this performance degradation is due to the arbitrary direction of adversarial attack. In other words, the augmentation module produces novel and challenging styles for the training; however, the styles are unrealistic and convey less information about real-world SC instances.

\begin{table}[]
\addtolength{\tabcolsep}{-2pt}
\small

\begin{center}
\resizebox{1\linewidth}{!}{
\begin{tabular}{lccccc}
\hline
\multirow{2}{*}{Method} & \multirow{2}{*}{SA} & \multicolumn{2}{c}{IJB-B} & \multicolumn{2}{c}{TinyFace} \\ \cline{3-6} 
                        &                                     & 1e-5        & 1e-4        & Rank-1        & Rank-5       \\ \hline
ArcFace                 & -                      & 86.05          & 93.40           & 62.70             & 65.85              \\

ArcFace                 & None-Targeted                      & 83.73          &93.34           & 61.50           & 66.33           \\
ArcFace                 & Targeted                          & \textbf{87.22}           & \textbf{94.33}           & \textbf{64.40}             & \textbf{68.53}  \\ \hline
\end{tabular}}
\vspace{5pt}
\caption{\small{Experiment on the impact of employing adversarial signal to compute novel style information}}\label{orthoimprove2} 
\end{center}
\end{table}

\subsection{Trainin Speed and Memory Consumption}
One significant benefit of the TSA module is the augmentation without needing image space manipulation. In Table \ref{speedmem}, we compare the proposed method's training speed and GPU memory usage with CFSM. Accordingly, by avoiding image space augmentation, we increased the training speed of the CFSM by almost 70\%. Furthermore, our model strives to dramatically reduce GPU memory consumption by removing the need for a generative module during the training.

\begin{table}[]
\begin{center}
\small

\resizebox{1\linewidth}{!}{
\begin{tabular}{lcccc}
\toprule
Method & Backbone & Batch Size & Mem $\downarrow$ & Speed $\uparrow$ \\ \hline
CFSM*   &  ResNet-50        &256        & 32  & 502   \\
TSA   &  ResNet-50        &256        & 21  & 872   \\ 
\bottomrule
\end{tabular}}
\vspace{5pt}
\caption{\small{Training speed and GPU memory consumption comparison between CFSM and the proposed method. Our proposal significantly enhances training efficiency and reduces GPU consumption compared to CFSM. ‘*’ denotes that we re-run the code due to the unavailability of the mentioned method trained on the MS1MV2 data in the official GitHub repository.}}\label{speedmem} 
\end{center}
\end{table}

\section{Conclusion}
This paper presented a simple yet effective training paradigm for face recognition. The proposed procedure,
mixes the feature statistics from the labeled and unlabeled face datasets to synthesize novel style information, which is inspired by the observation of the susceptibility of current face recognition models to the style information of the training set. Our method dynamically identifies the unrecognizable instances from unlabeled datasets to avoid synthesizing unrecognizable styles. This framework, allows the face recognition model to become exposed to novel and realistic styles during the training and improve its generalization. The efficacy of our method is evaluated through various experiments and evaluation across different benchmarks, including IJB-B, IJB-C, IJB-S, TinyFace, and SCFace.

{\small
\bibliographystyle{ieee}
\bibliography{egbib}

\begin{thebibliography}{10}\itemsep=-1pt

\bibitem{ahonen2008recognition}
T.~Ahonen, E.~Rahtu, V.~Ojansivu, and J.~Heikkila.
\newblock Recognition of blurred faces using local phase quantization.
\newblock In {\em 2008 19th international conference on pattern recognition}, pages 1--4. IEEE, 2008.

\bibitem{antoniou2017data}
A.~Antoniou, A.~Storkey, and H.~Edwards.
\newblock Data augmentation generative adversarial networks.
\newblock {\em arXiv preprint arXiv:1711.04340}, 2017.

\bibitem{chai2023recognizability}
J.~C.~L. Chai, T.-S. Ng, C.-Y. Low, J.~Park, and A.~B.~J. Teoh.
\newblock Recognizability embedding enhancement for very low-resolution face recognition and quality estimation.
\newblock In {\em Proceedings of the IEEE/CVF Conference on Computer Vision and Pattern Recognition}, pages 9957--9967, 2023.

\bibitem{chen2018fsrnet}
Y.~Chen, Y.~Tai, X.~Liu, C.~Shen, and J.~Yang.
\newblock Fsrnet: End-to-end learning face super-resolution with facial priors.
\newblock In {\em Proceedings of the IEEE conference on computer vision and pattern recognition}, pages 2492--2501, 2018.

\bibitem{saas}
Z.~Cheng, X.~Zhu, and S.~Gong.
\newblock Low-resolution face recognition.
\newblock In {\em Computer Vision--ACCV 2018: 14th Asian Conference on Computer Vision, Perth, Australia, December 2--6, 2018, Revised Selected Papers, Part III 14}, pages 605--621. Springer, 2019.

\bibitem{cheng2019low}
Z.~Cheng, X.~Zhu, and S.~Gong.
\newblock Low-resolution face recognition.
\newblock In {\em Computer Vision--ACCV 2018: 14th Asian Conference on Computer Vision, Perth, Australia, December 2--6, 2018, Revised Selected Papers, Part III 14}, pages 605--621. Springer, 2019.

\bibitem{dan2023homda}
J.~Dan, T.~Jin, H.~Chi, Y.~Shen, J.~Yu, and J.~Zhou.
\newblock Homda: High-order moment-based domain alignment for unsupervised domain adaptation.
\newblock {\em Knowledge-Based Systems}, 261:110205, 2023.

\bibitem{dan2023transface}
J.~Dan, Y.~Liu, H.~Xie, J.~Deng, H.~Xie, X.~Xie, and B.~Sun.
\newblock Transface: Calibrating transformer training for face recognition from a data-centric perspective.
\newblock In {\em Proceedings of the IEEE/CVF International Conference on Computer Vision}, pages 20642--20653, 2023.

\bibitem{deb2020advfaces}
D.~Deb, J.~Zhang, and A.~K. Jain.
\newblock Advfaces: Adversarial face synthesis.
\newblock In {\em 2020 IEEE International Joint Conference on Biometrics (IJCB)}, pages 1--10. IEEE, 2020.

\bibitem{deng2019arcface}
J.~Deng, J.~Guo, N.~Xue, and S.~Zafeiriou.
\newblock Arcface: Additive angular margin loss for deep face recognition.
\newblock In {\em Proceedings of the IEEE/CVF conference on computer vision and pattern recognition}, pages 4690--4699, 2019.

\bibitem{deng2023harnessing}
S.~Deng, Y.~Xiong, M.~Wang, W.~Xia, and S.~Soatto.
\newblock Harnessing unrecognizable faces for improving face recognition.
\newblock In {\em Proceedings of the IEEE/CVF Winter Conference on Applications of Computer Vision}, pages 3424--3433, 2023.

\bibitem{dong2018boosting}
Y.~Dong, F.~Liao, T.~Pang, H.~Su, J.~Zhu, X.~Hu, and J.~Li.
\newblock Boosting adversarial attacks with momentum.
\newblock In {\em Proceedings of the IEEE conference on computer vision and pattern recognition}, pages 9185--9193, 2018.

\bibitem{dong2019efficient}
Y.~Dong, H.~Su, B.~Wu, Z.~Li, W.~Liu, T.~Zhang, and J.~Zhu.
\newblock Efficient decision-based black-box adversarial attacks on face recognition.
\newblock In {\em Proceedings of the IEEE/CVF Conference on Computer Vision and Pattern Recognition}, pages 7714--7722, 2019.

\bibitem{fan2021adversarially}
X.~Fan, Q.~Wang, J.~Ke, F.~Yang, B.~Gong, and M.~Zhou.
\newblock Adversarially adaptive normalization for single domain generalization.
\newblock In {\em Proceedings of the IEEE/CVF Conference on Computer Vision and Pattern Recognition}, pages 8208--8217, 2021.

\bibitem{ge2020efficient}
S.~Ge, S.~Zhao, C.~Li, Y.~Zhang, and J.~Li.
\newblock Efficient low-resolution face recognition via bridge distillation.
\newblock {\em IEEE Transactions on Image Processing}, 29:6898--6908, 2020.

\bibitem{goodfellow2014explaining}
I.~J. Goodfellow, J.~Shlens, and C.~Szegedy.
\newblock Explaining and harnessing adversarial examples.
\newblock {\em arXiv preprint arXiv:1412.6572}, 2014.

\bibitem{grgic2011scface}
M.~Grgic, K.~Delac, and S.~Grgic.
\newblock Scface--surveillance cameras face database.
\newblock {\em Multimedia tools and applications}, 51:863--879, 2011.

\bibitem{guo2021sample}
J.~Guo, J.~Deng, A.~Lattas, and S.~Zafeiriou.
\newblock Sample and computation redistribution for efficient face detection.
\newblock {\em arXiv preprint arXiv:2105.04714}, 2021.

\bibitem{guo2016ms}
Y.~Guo, L.~Zhang, Y.~Hu, X.~He, and J.~Gao.
\newblock Ms-celeb-1m: A dataset and benchmark for large-scale face recognition.
\newblock In {\em Computer Vision--ECCV 2016: 14th European Conference, Amsterdam, The Netherlands, October 11-14, 2016, Proceedings, Part III 14}, pages 87--102. Springer, 2016.

\bibitem{he2020momentum}
K.~He, H.~Fan, Y.~Wu, S.~Xie, and R.~Girshick.
\newblock Momentum contrast for unsupervised visual representation learning.
\newblock In {\em Proceedings of the IEEE/CVF conference on computer vision and pattern recognition}, pages 9729--9738, 2020.

\bibitem{hoffman2018cycada}
J.~Hoffman, E.~Tzeng, T.~Park, J.-Y. Zhu, P.~Isola, K.~Saenko, A.~Efros, and T.~Darrell.
\newblock Cycada: Cycle-consistent adversarial domain adaptation.
\newblock In {\em International conference on machine learning}, pages 1989--1998. Pmlr, 2018.

\bibitem{huang2008labeled}
G.~B. Huang, M.~Mattar, T.~Berg, and E.~Learned-Miller.
\newblock Labeled faces in the wild: A database forstudying face recognition in unconstrained environments.
\newblock In {\em Workshop on faces in'Real-Life'Images: detection, alignment, and recognition}, 2008.

\bibitem{huang2017beyond}
R.~Huang, S.~Zhang, T.~Li, and R.~He.
\newblock Beyond face rotation: Global and local perception gan for photorealistic and identity preserving frontal view synthesis.
\newblock In {\em Proceedings of the IEEE international conference on computer vision}, pages 2439--2448, 2017.

\bibitem{huang2017arbitrary}
X.~Huang and S.~Belongie.
\newblock Arbitrary style transfer in real-time with adaptive instance normalization.
\newblock In {\em Proceedings of the IEEE international conference on computer vision}, pages 1501--1510, 2017.

\bibitem{jaynes1957information}
E.~T. Jaynes.
\newblock Information theory and statistical mechanics.
\newblock {\em Physical review}, 106(4):620, 1957.

\bibitem{jia2022adv}
S.~Jia, B.~Yin, T.~Yao, S.~Ding, C.~Shen, X.~Yang, and C.~Ma.
\newblock Adv-attribute: Inconspicuous and transferable adversarial attack on face recognition.
\newblock {\em arXiv preprint arXiv:2210.06871}, 2022.

\bibitem{kalarot2020component}
R.~Kalarot, T.~Li, and F.~Porikli.
\newblock Component attention guided face super-resolution network: Cagface.
\newblock In {\em Proceedings of the IEEE/CVF winter conference on applications of computer vision}, pages 370--380, 2020.

\bibitem{1}
N.~D. Kalka, B.~Maze, J.~A. Duncan, K.~O’Connor, S.~Elliott, K.~Hebert, J.~Bryan, and A.~K. Jain.
\newblock Ijb--s: Iarpa janus surveillance video benchmark.
\newblock In {\em 2018 IEEE 9th international conference on biometrics theory, applications and systems (BTAS)}, pages 1--9. IEEE, 2018.

\bibitem{kalka2018ijb}
N.~D. Kalka, B.~Maze, J.~A. Duncan, K.~O’Connor, S.~Elliott, K.~Hebert, J.~Bryan, and A.~K. Jain.
\newblock Ijb--s: Iarpa janus surveillance video benchmark.
\newblock In {\em 2018 IEEE 9th international conference on biometrics theory, applications and systems (BTAS)}, pages 1--9. IEEE, 2018.

\bibitem{kang2022style}
J.~Kang, S.~Lee, N.~Kim, and S.~Kwak.
\newblock Style neophile: Constantly seeking novel styles for domain generalization.
\newblock In {\em Proceedings of the IEEE/CVF Conference on Computer Vision and Pattern Recognition}, pages 7130--7140, 2022.

\bibitem{kim2019progressive}
D.~Kim, M.~Kim, G.~Kwon, and D.-S. Kim.
\newblock Progressive face super-resolution via attention to facial landmark.
\newblock {\em arXiv preprint arXiv:1908.08239}, 2019.

\bibitem{kim2022adaface}
M.~Kim, A.~K. Jain, and X.~Liu.
\newblock Adaface: Quality adaptive margin for face recognition.
\newblock In {\em Proceedings of the IEEE/CVF Conference on Computer Vision and Pattern Recognition}, pages 18750--18759, 2022.

\bibitem{komkov2021advhat}
S.~Komkov and A.~Petiushko.
\newblock Advhat: Real-world adversarial attack on arcface face id system.
\newblock In {\em 2020 25th International Conference on Pattern Recognition (ICPR)}, pages 819--826. IEEE, 2021.

\bibitem{kullback1997information}
S.~Kullback.
\newblock {\em Information theory and statistics}.
\newblock Courier Corporation, 1997.

\bibitem{li2023rethinking}
J.~Li, Z.~Guo, H.~Li, S.~Han, J.-w. Baek, M.~Yang, R.~Yang, and S.~Suh.
\newblock Rethinking feature-based knowledge distillation for face recognition.
\newblock In {\em Proceedings of the IEEE/CVF Conference on Computer Vision and Pattern Recognition}, pages 20156--20165, 2023.

\bibitem{liu2022controllable}
F.~Liu, M.~Kim, A.~Jain, and X.~Liu.
\newblock Controllable and guided face synthesis for unconstrained face recognition.
\newblock In {\em Computer Vision--ECCV 2022: 17th European Conference, Tel Aviv, Israel, October 23--27, 2022, Proceedings, Part XII}, pages 701--719. Springer, 2022.

\bibitem{lucas1990exponentially}
J.~M. Lucas and M.~S. Saccucci.
\newblock Exponentially weighted moving average control schemes: properties and enhancements.
\newblock {\em Technometrics}, 32(1):1--12, 1990.

\bibitem{ma2020deep}
C.~Ma, Z.~Jiang, Y.~Rao, J.~Lu, and J.~Zhou.
\newblock Deep face super-resolution with iterative collaboration between attentive recovery and landmark estimation.
\newblock In {\em Proceedings of the IEEE/CVF conference on computer vision and pattern recognition}, pages 5569--5578, 2020.

\bibitem{madry2017towards}
A.~Madry, A.~Makelov, L.~Schmidt, D.~Tsipras, and A.~Vladu.
\newblock Towards deep learning models resistant to adversarial attacks.
\newblock {\em arXiv preprint arXiv:1706.06083}, 2017.

\bibitem{maze2018iarpa}
B.~Maze, J.~Adams, J.~A. Duncan, N.~Kalka, T.~Miller, C.~Otto, A.~K. Jain, W.~T. Niggel, J.~Anderson, J.~Cheney, et~al.
\newblock Iarpa janus benchmark-c: Face dataset and protocol.
\newblock In {\em 2018 International Conference on Biometrics (ICB)}, pages 158--165. IEEE, 2018.

\bibitem{meishvili2020learning}
G.~Meishvili, S.~Jenni, and P.~Favaro.
\newblock Learning to have an ear for face super-resolution.
\newblock In {\em Proceedings of the IEEE/CVF conference on computer vision and pattern recognition}, pages 1364--1374, 2020.

\bibitem{pal1991entropy}
N.~R. Pal and S.~K. Pal.
\newblock Entropy: A new definition and its applications.
\newblock {\em IEEE transactions on systems, man, and cybernetics}, 21(5):1260--1270, 1991.

\bibitem{qiao2021uncertainty}
F.~Qiao and X.~Peng.
\newblock Uncertainty-guided model generalization to unseen domains.
\newblock In {\em Proceedings of the IEEE/CVF conference on computer vision and pattern recognition}, pages 6790--6800, 2021.

\bibitem{qiao2020learning}
F.~Qiao, L.~Zhao, and X.~Peng.
\newblock Learning to learn single domain generalization.
\newblock In {\em Proceedings of the IEEE/CVF Conference on Computer Vision and Pattern Recognition}, pages 12556--12565, 2020.

\bibitem{robbins2022effect}
W.~Robbins and T.~E. Boult.
\newblock On the effect of atmospheric turbulence in the feature space of deep face recognition.
\newblock In {\em Proceedings of the IEEE/CVF Conference on Computer Vision and Pattern Recognition}, pages 1618--1626, 2022.

\bibitem{sankaranarayanan2018generate}
S.~Sankaranarayanan, Y.~Balaji, C.~D. Castillo, and R.~Chellappa.
\newblock Generate to adapt: Aligning domains using generative adversarial networks.
\newblock In {\em Proceedings of the IEEE conference on computer vision and pattern recognition}, pages 8503--8512, 2018.

\bibitem{shannon2001mathematical}
C.~E. Shannon.
\newblock A mathematical theory of communication.
\newblock {\em ACM SIGMOBILE mobile computing and communications review}, 5(1):3--55, 2001.

\bibitem{sharif2016accessorize}
M.~Sharif, S.~Bhagavatula, L.~Bauer, and M.~K. Reiter.
\newblock Accessorize to a crime: Real and stealthy attacks on state-of-the-art face recognition.
\newblock In {\em Proceedings of the 2016 acm sigsac conference on computer and communications security}, pages 1528--1540, 2016.

\bibitem{shi2019probabilistic}
Y.~Shi and A.~K. Jain.
\newblock Probabilistic face embeddings.
\newblock In {\em Proceedings of the IEEE/CVF International Conference on Computer Vision}, pages 6902--6911, 2019.

\bibitem{shi2021boosting}
Y.~Shi and A.~K. Jain.
\newblock Boosting unconstrained face recognition with auxiliary unlabeled data.
\newblock In {\em Proceedings of the IEEE/CVF Conference on Computer Vision and Pattern Recognition}, pages 2795--2804, 2021.

\bibitem{shin2022teaching}
S.~Shin, J.~Lee, J.~Lee, Y.~Yu, and K.~Lee.
\newblock Teaching where to look: Attention similarity knowledge distillation for low resolution face recognition.
\newblock In {\em Computer Vision--ECCV 2022: 17th European Conference, Tel Aviv, Israel, October 23--27, 2022, Proceedings, Part XII}, pages 631--647. Springer Nature Switzerland Cham, 2022.

\bibitem{singh2019dual}
M.~Singh, S.~Nagpal, R.~Singh, and M.~Vatsa.
\newblock Dual directed capsule network for very low resolution image recognition.
\newblock In {\em Proceedings of the IEEE/CVF International Conference on Computer Vision}, pages 340--349, 2019.

\bibitem{sun2021mae}
Z.~Sun, M.~Lin, X.~Sun, Z.~Tan, H.~Li, and R.~Jin.
\newblock Mae-det: Revisiting maximum entropy principle in zero-shot nas for efficient object detection.
\newblock {\em arXiv preprint arXiv:2111.13336}, 2021.

\bibitem{terhorst2023qmagface}
P.~Terh{\"o}rst, M.~Ihlefeld, M.~Huber, N.~Damer, F.~Kirchbuchner, K.~Raja, and A.~Kuijper.
\newblock Qmagface: Simple and accurate quality-aware face recognition.
\newblock In {\em Proceedings of the IEEE/CVF Winter Conference on Applications of Computer Vision}, pages 3484--3494, 2023.

\bibitem{thomas1991elements}
J.~A. Thomas.
\newblock Elements of information theory, 1991.

\bibitem{ulyanov2016instance}
D.~Ulyanov, A.~Vedaldi, and V.~Lempitsky.
\newblock Instance normalization: The missing ingredient for fast stylization.
\newblock {\em arXiv preprint arXiv:1607.08022}, 2016.

\bibitem{volpi2018generalizing}
R.~Volpi, H.~Namkoong, O.~Sener, J.~C. Duchi, V.~Murino, and S.~Savarese.
\newblock Generalizing to unseen domains via adversarial data augmentation.
\newblock {\em Advances in neural information processing systems}, 31, 2018.

\bibitem{wang2018cosface}
H.~Wang, Y.~Wang, Z.~Zhou, X.~Ji, D.~Gong, J.~Zhou, Z.~Li, and W.~Liu.
\newblock Cosface: Large margin cosine loss for deep face recognition.
\newblock In {\em Proceedings of the IEEE conference on computer vision and pattern recognition}, pages 5265--5274, 2018.

\bibitem{wang2019improved}
M.~Wang, R.~Liu, N.~Hajime, A.~Narishige, H.~Uchida, and T.~Matsunami.
\newblock Improved knowledge distillation for training fast low resolution face recognition model.
\newblock In {\em Proceedings of the IEEE/CVF International Conference on Computer Vision Workshops}, pages 0--0, 2019.

\bibitem{wen2021sphereface2}
Y.~Wen, W.~Liu, A.~Weller, B.~Raj, and R.~Singh.
\newblock Sphereface2: Binary classification is all you need for deep face recognition.
\newblock {\em arXiv preprint arXiv:2108.01513}, 2021.

\bibitem{whitelam2017iarpa}
C.~Whitelam, E.~Taborsky, A.~Blanton, B.~Maze, J.~Adams, T.~Miller, N.~Kalka, A.~K. Jain, J.~A. Duncan, K.~Allen, et~al.
\newblock Iarpa janus benchmark-b face dataset.
\newblock In {\em proceedings of the IEEE Conference on Computer Vision and Pattern Recognition workshops}, pages 90--98, 2017.

\bibitem{yang2014single}
C.-Y. Yang, C.~Ma, and M.-H. Yang.
\newblock Single-image super-resolution: A benchmark.
\newblock In {\em Computer Vision--ECCV 2014: 13th European Conference, Zurich, Switzerland, September 6-12, 2014, Proceedings, Part IV 13}, pages 372--386. Springer, 2014.

\bibitem{yang2022adversarial}
K.~Yang, Y.~Sun, J.~Su, F.~He, X.~Tian, F.~Huang, T.~Zhou, and D.~Tao.
\newblock Adversarial auto-augment with label preservation: A representation learning principle guided approach.
\newblock {\em Advances in Neural Information Processing Systems}, 35:22035--22048, 2022.

\bibitem{yang2016wider}
S.~Yang, P.~Luo, C.-C. Loy, and X.~Tang.
\newblock Wider face: A face detection benchmark.
\newblock In {\em Proceedings of the IEEE conference on computer vision and pattern recognition}, pages 5525--5533, 2016.

\bibitem{yang2019fsa}
T.-Y. Yang, Y.-T. Chen, Y.-Y. Lin, and Y.-Y. Chuang.
\newblock Fsa-net: Learning fine-grained structure aggregation for head pose estimation from a single image.
\newblock In {\em Proceedings of the IEEE/CVF conference on computer vision and pattern recognition}, pages 1087--1096, 2019.

\bibitem{yang2019deep}
W.~Yang, X.~Zhang, Y.~Tian, W.~Wang, J.-H. Xue, and Q.~Liao.
\newblock Deep learning for single image super-resolution: A brief review.
\newblock {\em IEEE Transactions on Multimedia}, 21(12):3106--3121, 2019.

\bibitem{yin2021adv}
B.~Yin, W.~Wang, T.~Yao, J.~Guo, Z.~Kong, S.~Ding, J.~Li, and C.~Liu.
\newblock Adv-makeup: A new imperceptible and transferable attack on face recognition.
\newblock {\em arXiv preprint arXiv:2105.03162}, 2021.

\bibitem{yu2018face}
X.~Yu, B.~Fernando, B.~Ghanem, F.~Porikli, and R.~Hartley.
\newblock Face super-resolution guided by facial component heatmaps.
\newblock In {\em Proceedings of the European conference on computer vision (ECCV)}, pages 217--233, 2018.

\bibitem{zangeneh2020low}
E.~Zangeneh, M.~Rahmati, and Y.~Mohsenzadeh.
\newblock Low resolution face recognition using a two-branch deep convolutional neural network architecture.
\newblock {\em Expert Systems with Applications}, 139:112854, 2020.

\bibitem{zhang2022exact}
Y.~Zhang, M.~Li, R.~Li, K.~Jia, and L.~Zhang.
\newblock Exact feature distribution matching for arbitrary style transfer and domain generalization.
\newblock In {\em Proceedings of the IEEE/CVF Conference on Computer Vision and Pattern Recognition}, pages 8035--8045, 2022.

\bibitem{zhong2022adversarial}
Z.~Zhong, Y.~Zhao, G.~H. Lee, and N.~Sebe.
\newblock Adversarial style augmentation for domain generalized urban-scene segmentation.
\newblock {\em Advances in Neural Information Processing Systems}, 35:338--350, 2022.

\bibitem{zhou2021domain}
K.~Zhou, Y.~Yang, Y.~Qiao, and T.~Xiang.
\newblock Domain generalization with mixstyle.
\newblock {\em arXiv preprint arXiv:2104.02008}, 2021.

\bibitem{zhu2017unpaired}
J.-Y. Zhu, T.~Park, P.~Isola, and A.~A. Efros.
\newblock Unpaired image-to-image translation using cycle-consistent adversarial networks.
\newblock In {\em Proceedings of the IEEE international conference on computer vision}, pages 2223--2232, 2017.

\end{thebibliography}
}

\end{document}